\documentclass[11pt,a4paper]{article}
\usepackage[hyperref]{emnlp-ijcnlp-2019}
\usepackage{times}
\usepackage{latexsym}

\usepackage{url}

\usepackage{microtype}
\usepackage{graphicx}
\usepackage{subfigure}
\usepackage{amsmath}
\usepackage{amsfonts}
\usepackage{booktabs} 

\aclfinalcopy 

\title{Style Transfer for Texts: Retrain, Report Errors, Compare with Rewrites}

\author{Alexey Tikhonov\thanks{Equal contribution} \\
  Yandex \\
  Berlin, Germany \\
  \texttt{altsoph@gmail.com} \\\And
  Viacheslav Shibaev\footnotemark[1]  \\
  Ural Federal University \\
  Ekaterinburg, Russia \\\And
  Aleksander Nagaev  \\
  Ural Federal University/Sberbank \\
  Ekaterinburg, Russia \\\AND
  Aigul Nugmanova \\
  Speech Technology Center \\
  St. Petersburg, Russia \\\And
  Ivan P. Yamshchikov\footnotemark[1]  \\
  Max Planck Institute \\
  for Mathematics in the Sciences \\
  Leipzig, Germany \\
  \texttt{ivan@yamshchikov.info} \\}

\date{}

\begin{document}
\maketitle
\begin{abstract}
This paper shows that standard assessment methodology for style transfer has several significant problems. First, the standard metrics for style accuracy and semantics preservation vary significantly on different re-runs. Therefore one has to report error margins for the obtained results. Second, starting with certain values of bilingual evaluation understudy (BLEU) between input and output and accuracy of the sentiment transfer the optimization of these two standard metrics diverge from the intuitive goal of the style transfer task. Finally, due to the nature of the task itself, there is a specific dependence between these two metrics that could be easily manipulated. Under these circumstances, we suggest taking BLEU between input and human-written reformulations into consideration for benchmarks. We also propose three new architectures that outperform state of the art in terms of this metric.
\end{abstract}

\section{Introduction}
\label{intro}

Deep generative models attract a lot of attention in recent years \cite{hu17}. Such methods as variational autoencoders \cite{kingma13} or generative adversarial networks \cite{goodfellow} are successfully applied to a variety of machine vision problems including image generation \cite{Radford}, learning interpretable image representations \cite{Chen} and style transfer for images \cite{Gatys}. However, natural language generation is more challenging due to many reasons, such as the discrete nature of textual information \cite{hylsx}, the absence of local information continuity and non-smooth disentangled representations \cite{bowman}. Due to these difficulties, text generation is mostly limited to specific narrow applications and is usually working in supervised settings. 

Content and style are deeply fused in natural language, but style transfer for texts is often addressed in the context of disentangled latent representations \cite{hylsx, Shen, fu2, john18, romanov18, tian18}. Intuitive understanding of this problem is apparent: if an input text has some attribute $A$, a system generates new text similar to the input on a given set of attributes with only one attribute $A$ changed to the target attribute $\tilde{A}$. In the majority of previous works, style transfer is obtained through an encoder-decoder architecture with one or multiple style discriminators to learn disentangled representations. The encoder takes a sentence as an input and generates a style-independent content representation. The decoder then takes the content representation and the target style representation to generate the transformed sentence. In \cite{subramanian18} authors question the quality and usability of the disentangled representations for texts and suggest an end-to-end approach to style transfer similar to an end-to-end machine translation. 

\begin{figure}[ht]
\begin{center}
\centerline{\includegraphics[width=\columnwidth]{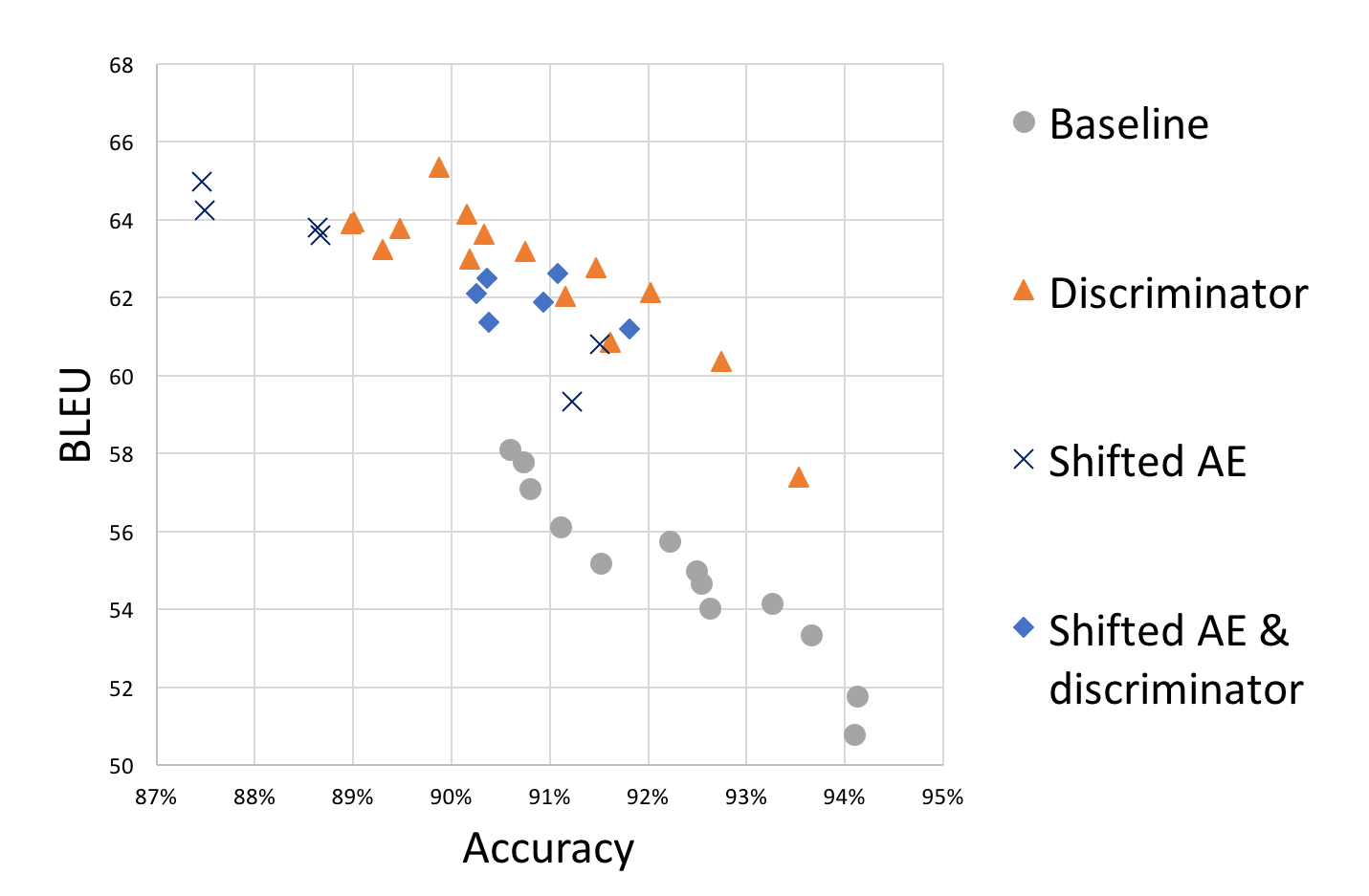}}
\caption{Test results of multiple runs for four different architectures retrained several times from scratch. In-depth description of the architectures can be found in Section \ref{sec:st}.}
\label{pic:ex}
\end{center}
\end{figure}

Contribution of this paper is three-fold: 1) we show that different style transfer architectures have varying results on test and that reporting error margins for various training re-runs of the same model is especially important for adequate assessment of the models accuracy, see Figure \ref{pic:ex}; 2) we show that BLEU \cite{Papineni} between input and output and accuracy of style transfer measured in terms of the accuracy of a pre-trained external style classifier can be manipulated and naturally diverge from the intuitive goal of the style transfer task starting from a certain threshold; 3) new architectures that perform style transfer using improved latent representations are shown to outperform state of the art in terms of BLEU between output and human-written reformulations.

\section{Related Work}
\label{sec:rw}
 
Style of a text is a very general notion that is hard to define in rigorous terms \cite{Xu3}. However, the style of a text can be characterized quantitatively \cite{Hughes}; stylized texts could be generated if a system is trained on a dataset of stylistically similar texts \cite{Potash}; and author-style could be learned end-to-end \cite{TY, TYwilde, Vechtomova}. A majority of recent works on style transfer focus on the sentiment of text and use it as a target attribute. For example, in \cite{li, Kabbara, Xu2} estimate the quality of the style transfer with binary sentiment classifier trained on the corpora further used for the training of the style-transfer system. \cite{Ficler} and especially \cite{fu2} generalize this ad-hoc approach defining a style as a set of arbitrary quantitively measurable categorial or continuous parameters. Such parameters could include the \textit{'style of the time'} \cite{Hughes}, author-specific attributes (see \cite{xu} or \cite{Jhamtani} on 'shakespearization'), politeness \cite{Sennrich}, formality of speech \cite{Rao}, and gender or even political slant \cite{Prabhumoye}.

A significant challenge associated with narrowly defined style-transfer problems is that finding a good solution for one aspect of a style does not guarantee that you can use the same solution for a different aspect of it. For example,  \citet{guu} build a generative model for sentiment transfer with a retrieve-edit approach. In \cite{li} a delete-retrieve model shows good results for sentiment transfer. However, it is hard to imagine that these retrieval approaches could be used, say, for the style of the time or formality, since in these cases the system is often expected to paraphrase a given sentence to achieve the target style. 

In \cite{hylsx} the authors propose a more general approach to the controlled text generation combining variational autoencoder (VAE) with an extended wake-sleep mechanism in which the sleep procedure updates both the generator and external classifier that assesses generated samples and feedbacks learning signals to the generator. Authors had concatenated labels for style with the text representation of the encoder and used this vector with "hard-coded" information about the sentiment of the output as the input of the decoder. This approach seems promising, and some other papers either extend it or use similar ideas. \citet{Shen} applied a GAN to align the hidden representations of sentences from two corpora using an adversarial loss to decompose information about the form. In \cite{zhao} model learns a smooth code space and can be used as a discrete GAN with the ability to generate coherent discrete outputs from continuous samples. Authors use two different generators for two different styles. In \cite{fu2} an adversarial network is used to make sure that the output of the encoder does not have style representation. \cite{hylsx} also uses an adversarial component that ensures there is no stylistic information within the representation. \citet{fu2} do not use a dedicated component that controls the semantic component of the latent representation.  Such a component is proposed by \citet{john18} who demonstrate that decomposition of style and content could be improved with an auxiliary multi-task for label prediction and adversarial objective for bag-of-words prediction. \citet{romanov18} also introduces a dedicated component to control semantic aspects of latent representations and an adversarial-motivational training that includes a special motivational loss to encourage a better decomposition. Speaking about preservation of semantics one also has to mention works on paraphrase systems, see, for example \cite{para1, para2, para3}. The methodology described in this paper could be extended to paraphrasing systems in terms of semantic preservation measurement, however, this is the matter of future work.

\citet{subramanian18} state that learning a latent representation, which is independent of the attributes specifying its style, is rarely attainable. There are other works on style transfer that are based on the ideas of neural machine translation with \cite{Carlson} and without parallel corpora \cite{zhang18} in line with \cite{Lample} and \cite{Artetxe}.

\begin{figure}[ht]
\begin{center}
\centerline{\includegraphics[width=\columnwidth]{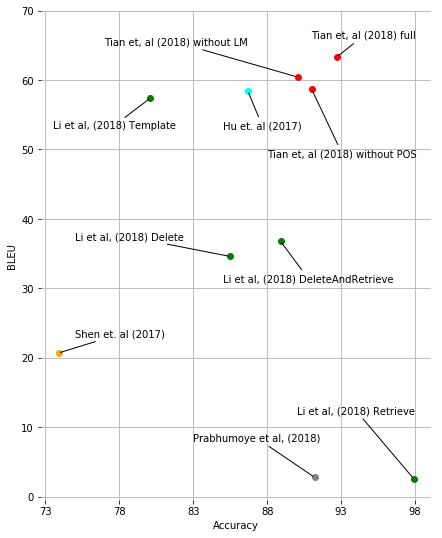}}
\caption{Overview of the self-reported results for sentiment transfer on Yelp! reviews. Results of \cite{romanov18} are not displayed due to the absence of self-reported BLEU scores. Later in the paper we show that on different reruns BLEU and accuracy can vary from these self-reported single results.}
\label{pic:il}
\end{center}
\end{figure}

It is important to underline here that majority of the papers dedicated to style transfer for texts treat sentiment of a sentence as a stylistic rather than semantic attribute despite particular concerns \cite{TYwrong}. It is also crucial to mention that in line with \cite{fu2} majority of the state of the art methods for style transfer use an external pre-trained classifier to measure the accuracy of the style transfer. BLEU computes the harmonic mean of precision of exact matching n-grams between a reference and a target sentence across the corpus. It is not sensitive to minute changes, but BLEU between input and output is often used as the coarse measure of the semantics preservation. For the corpora that have human written reformulations, BLEU between the output of the model and human text is used. These metrics are used alongside with a handful of others such as PINC (Paraphrase In N-gram Changes) score \cite{Carlson}, POS distance \cite{tian18}, language fluency \cite{john18}, etc. Figure \ref{pic:il} shows self-reported results of different models in terms of two most frequently measured performance metrics, namely, BLEU and Accuracy of the style transfer. 

This paper focuses on Yelp! reviews dataset\footnote{https://www.yelp.com/dataset} that was lately enhanced with human written reformulations by \cite{li}. These are Yelp! reviews, where each short English review of a place is labeled as a negative or as a positive once. This paper studies three metrics that are most common in the field at the moment and questions to which extent can they be used for the performance assessment. These metrics are the accuracy of an external style classifier that is trained to measure the accuracy of the style transfer, BLEU between input and output of a system, and BLEU between output and human-written texts.


\section{Style transfer}
\label{sec:st}
\begin{figure}[ht]
\begin{center}
\centerline{\includegraphics[width=\columnwidth]{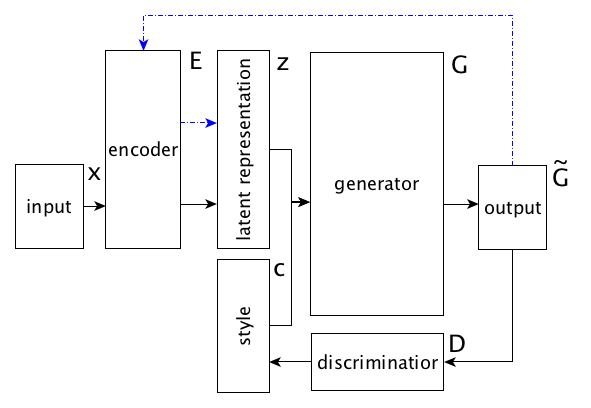}}
\caption{The generative model, where style is a structured code targeting sentence attributes to control.
Blue dashed arrows denote the proposed independence constraint of latent representation and controlled attribute, see \cite{hylsx} for the details.}
\label{pic:hu}
\end{center}
\end{figure}

In this work we experiment with extensions of a model, described in \cite{hylsx}, using Texar \cite{hutexar} framework. To generate plausible sentences with specific semantic and stylistic features every sentence is conditioned on a representation vector $z$ which is concatenated with a particular code $c$ that specifies desired attribute, see Figure \ref{pic:hu}. Under notation introduced in \cite{hylsx} the base autoencoder (AE) includes a conditional probabilistic encoder $E$ defined with parameters $\theta_E$ to infer the latent representation $z$ given input $x$
$$z \sim E(x) = q_{E}(z,c|x).$$
Generator $G$ defined with parameters  $\theta_G$ is a GRU-RNN for generating and output $\hat{x}$ defined as a sequence of tokens $\hat{x} = {\hat{x}_1, ..., \hat{x}_T}$ conditioned on the latent representation $z$ and a stylistic component $c$ that are concatenated and give rise to a generative distribution
$$\hat{x} \sim G(z,c) = p_G(\hat{x}|z, c).$$
These encoder and generator form an AE with the following loss
\begin{equation}
\label{eq:lossae}
\mathcal{L}_{ae} (\theta_G, \theta_E; x,c) = - \mathbb{E}_{q_{E} (z,c|x)} \left[ \log q_G (x|z, c) \right].
\end{equation}

This standard reconstruction loss that drives the generator to produce realistic sentences is combined with two additional losses. The first discriminator provides extra learning signals which enforce the generator to produce coherent attributes that match the structured code in $c$. Since it is impossible to propagate gradients from the discriminator through the discrete sample $\hat{x}$, we use a deterministic continuous approximation a "soft" generated sentence, denoted as $\tilde{G} = \tilde{G}_\tau (z, c)$ with "temperature" $\tau$ set to $\tau \rightarrow 0$ as training proceeds. The resulting “soft” generated sentence is fed into the discriminator to measure the fitness to the target attribute, leading to the following loss
\begin{equation}
\mathcal{L}_{c} (\theta_G, \theta_E; x) = -\mathbb{E}_{q_{E} (z,c|x)}  \left[ \log q_D (c | \tilde{G}) \right]. \label{eq:lossc}
\end{equation}

Finally, under the assumption that each structured attribute of generated sentences is controlled through the corresponding code in $c$ and is independent from $z$ one would like to control that other not explicitly modelled attributes do not entangle with  $c$. This is addressed by the  dedicated loss
\begin{equation}
\label{eq:lossz}
\mathcal{L}_{z} (\theta_G; x) = - \mathbb{E}_{q_{E} (z,c|x) q_{D} (c|x)} \left[ \log q_E (z | \tilde{G}) \right].
\end{equation}
The training objective for the baseline, shown in Figure \ref{pic:hu}, is therefore a sum of the losses from Equations (\ref{eq:lossae}) -- (\ref{eq:lossz}) defined as
\begin{equation}
\label{eq:genhu}
min_{\theta_G} \mathcal{L}_{baseline} = \mathcal{L}_{ae} + \lambda_c \mathcal{L}_{c} + \lambda_z \mathcal{L}_{z},
\end{equation}
where $\lambda_c$ and $\lambda_z$ are balancing parameters.

\begin{figure}[ht]
\begin{center}
\centerline{\includegraphics[width=\columnwidth]{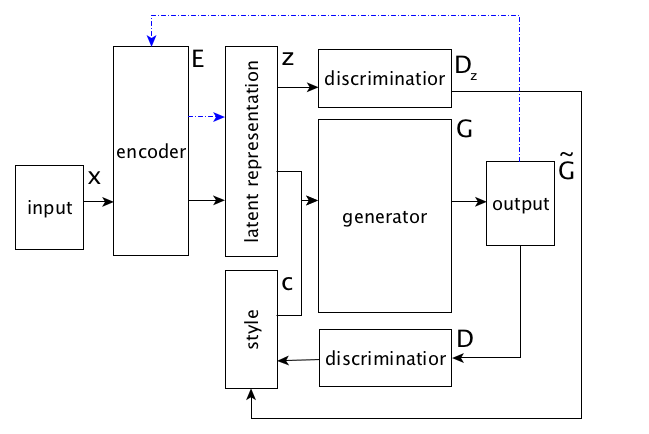}}
\caption{The generative model with dedicated discriminator introduced to ensure that semantic part of the latent representation does not have information on the style of the text.}
\label{pic:d}
\end{center}
\end{figure}

Let us propose two further extensions of this baseline architecture. To improve reproducibility of the research the code of the studied models is open\footnote{https://github.com/VAShibaev/text\_style\_transfer}. Both extensions aim to improve the quality of information decomposition within the latent representation. In the first one, shown in Figure \ref{pic:d}, a special dedicated discriminator is added to the model to control that the latent representation does not contain stylistic information. The loss of this discriminator is defined as
\begin{equation}
\label{eq:lossdisc}
\mathcal{L}_{D_z} (\theta_G; x,c) = - \mathbb{E}_{q_{E} (z|x)} \left[ \log q_{D_z} (c | z) \right].
\end{equation}

Here a discriminator denoted as $D_z$ is trying to predict code $c$ using representation $z$. Combining the loss defined by Equation (\ref{eq:genhu}) with the adversarial component defined in Equation (\ref{eq:lossdisc}) the following learning objective is formed
\begin{equation}
\label{eq:gendz}
min_{\theta_G} \mathcal{L} = \mathcal{L}_{baseline} - \lambda_{D_z} \mathcal{L}_{Dz},
\end{equation}
where $\mathcal{L}_{baseline}$ is a sum defined in Equation (\ref{eq:genhu}), $\lambda_{D_z}$ is a balancing parameter.

\begin{figure}[ht]
\begin{center}
\centerline{\includegraphics[width=\columnwidth]{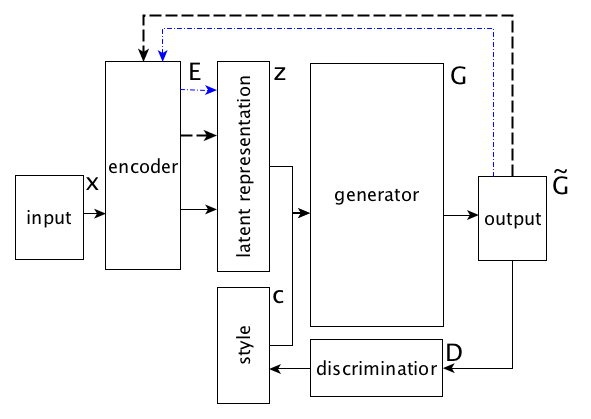}}
\caption{The generative model with a dedicated loss added to control that semantic representation of the output, when processed by the encoder, is close to the semantic representation of the input.}
\label{pic:sae}
\end{center}
\end{figure}

The second extension of the baseline architecture does not use an adversarial component $D_z$ that is trying to eradicate information on $c$ from component $z$. Instead, the system, shown in Figure \ref{pic:sae} feeds the "soft" generated sentence $\tilde{G}$ into encoder $E$ and checks how close is the representation $E(\tilde{G} )$ to the original representation $z = E(x)$ in terms of the cosine distance. We further refer to it as {\em shifted autoencoder} or SAE. Ideally, both $E(\tilde{G} (E(x), c))$ and $E(\tilde{G} (E(x), \bar{c}))$, where $\bar{c}$ denotes an inverse style code, should be both equal to $E(x)$\footnote{This notation is valid under the assumption that every stylistic attribute is a binary feature}. The loss of the shifted autoencoder is 
\begin{equation}
\label{eq:gesae}
min_{\theta_G} \mathcal{L} = \mathcal{L}_{baseline} +  \lambda_{cos} \mathcal{L}_{cos} +  \lambda_{cos^{-}} \mathcal{L}_{cos^{-}},
\end{equation}
where $\lambda_{cos}$ and $\lambda_{cos^{-}}$ are two balancing parameters, with two additional terms in the loss, namely, cosine distances between the softened output processed by the encoder and the encoded original input, defined as 
\begin{eqnarray}
\label{eq:cosloss}
\mathcal{L}_{cos} (x,c) = \cos \left( E(\tilde{G}(E(x), c)), E(x) \right),  \nonumber \\
\mathcal{L}_{cos^{-}} (x,c) = \cos \left( E(\tilde{G}(E(x), \bar{c})), E(x) \right).
\end{eqnarray}

\begin{figure}[ht]
\begin{center}
\centerline{\includegraphics[width=\columnwidth]{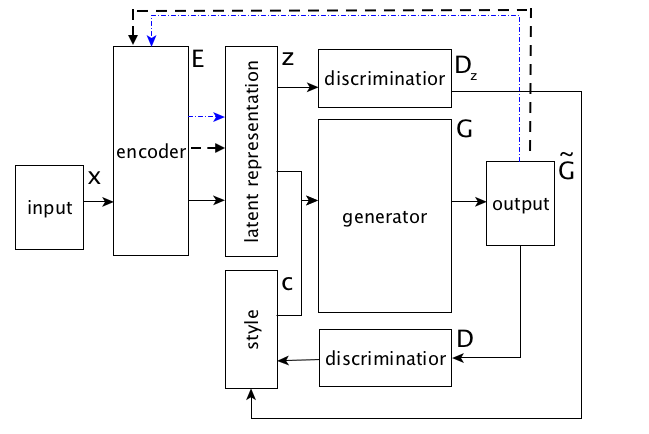}}
\caption{A combination of an additional discriminator used in Figure \ref{pic:d} with a shifted autoencoder shown in Figure \ref{pic:sae}}
\label{pic:combo}
\end{center}
\end{figure}

We also study a combination of both approaches described above, shown on Figure \ref{pic:combo}.

In Section \ref{sec:ex} we describe a series of experiments that we have carried out for these architectures using Yelp! reviews dataset.

\section{Experiments}
\label{sec:ex}
We have found that the baseline, as well as the proposed extensions,  have noisy outcomes, when retrained from scratch, see Figure \ref{pic:ex}. Most of the papers mentioned in Section \ref{sec:rw} measure the performance of the methods proposed for the sentiment transfer with two metrics: accuracy of the external sentiment classifier measured on test data, and BLEU between the input and output that is regarded as a coarse metric for semantic similarity.

In the first part of this section, we demonstrate that reporting error margins is essential for the performance assessment in terms that are prevalent in the field at the moment, i.e., BLEU between input and output and accuracy of the external sentiment classifier. In the second part, we also show that both of these two metrics after a certain threshold start to diverge from an intuitive goal of the style transfer and could be manipulated.

\subsection{Error margins matter}
\label{sec:no}

On Figure \ref{pic:ex} one can see that the outcomes for every single rerun differ significantly. Namely, accuracy can change up to 5 percentage points, whereas BLEU can vary up to 8 points. This variance can be partially explained with the stochasticity incurred due to sampling from the latent variables. However, we show that results for state of the art models sometimes end up within error margins from one another, so one has to report the margins to compare the results rigorously.   More importantly, one can see that there is an inherent trade-off between these two performance metrics. This trade-off is not only visible across models but is also present for the same retrained architecture. Therefore, improving one of the two metrics is not enough to confidently state that one system solves the style-transfer problem better than the other. One has to report error margins after several consecutive retrains and instead of comparing one of the two metrics has to talk about Pareto-like optimization that would show confident improvement of both.

\begin{figure}[ht]
\begin{center}
\centerline{\includegraphics[width=\columnwidth]{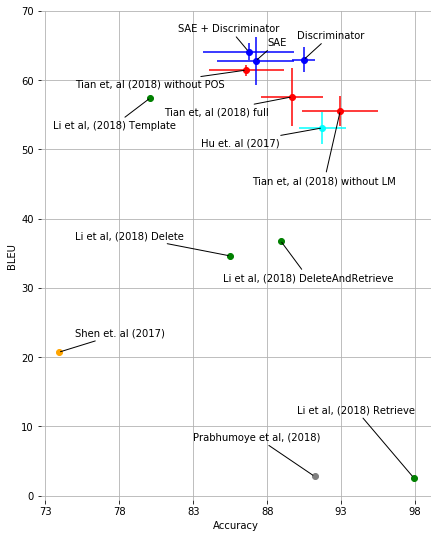}}
\caption{Overview of the self-reported results for sentiment transfer on Yelp! reviews alongside with the results for the baseline model \cite{hylsx}, architecture with additional discriminator, shifted autoencoder (SAE) with additional cosine losses, and a combination of these two architectures averaged after five re-trains alongside with architectures proposed by \cite{tian18} after five consecutive re-trains. Results of \cite{romanov18} are not displayed due to the absence of self-reported BLEU scores.}
\label{pic:ov}
\end{center}
\end{figure}

To put obtained results into perspective, we have retrained every model from scratch five times in a row. We have also retrained the models of \citet{tian18} five times since their code is published online. Figure \ref{pic:ov} shows the results of all models with error margins. It is also enhanced with other self-reported results on the same Yelp! review dataset for which no code was published. 

One can see that error margins of the models, for which several reruns could be performed, overlap significantly. In the next subsection, we carefully study BLEU and accuracy of the external classifier and discuss their aptness to measure style transfer performance.

\subsection{Delete, duplicate and conquer}
\label{sec:disc}

One can argue that as there is an inevitable entanglement between semantics and stylistics in natural language, there is also an apparent entanglement between BLEU of input and output and accuracy estimation of the style. Indeed, the output that copies input gives maximal BLEU yet clearly fails in terms of the style transfer. On the other hand, a wholly rephrased sentence could provide a low BLEU between input and output but high accuracy. These two issues are not problematic when both BLEU between input and output and accuracy of the transfer are relatively low. However, since style transfer methods have significantly evolved in recent years, some state of the art methods are now sensitive to these issues. The trade-off between these two metrics can be seen in Figure \ref{pic:ex} as well as in Figure \ref{pic:ov}.

As we have mentioned above, the accuracy of an external classifier and BLEU between output and input are the most widely used methods to assess the performance of style transfer at this moment. However, both of these metrics can be manipulated in a relatively simple manner. One can extend the generative architecture with internal pre-trained classifier of style and then perform the following heuristic procedure:
\begin{itemize}
    \item measure the style accuracy on the output for a given batch;
    \item choose the sentences that style classifier labels as incorrect;
    \item replace them with duplicates of sentences from the given batch that have correct style according to the internal classifier and show the highest BLEU with given inputs.
\end{itemize}

\begin{figure}[ht]
\begin{center}
\centerline{\includegraphics[width=\columnwidth]{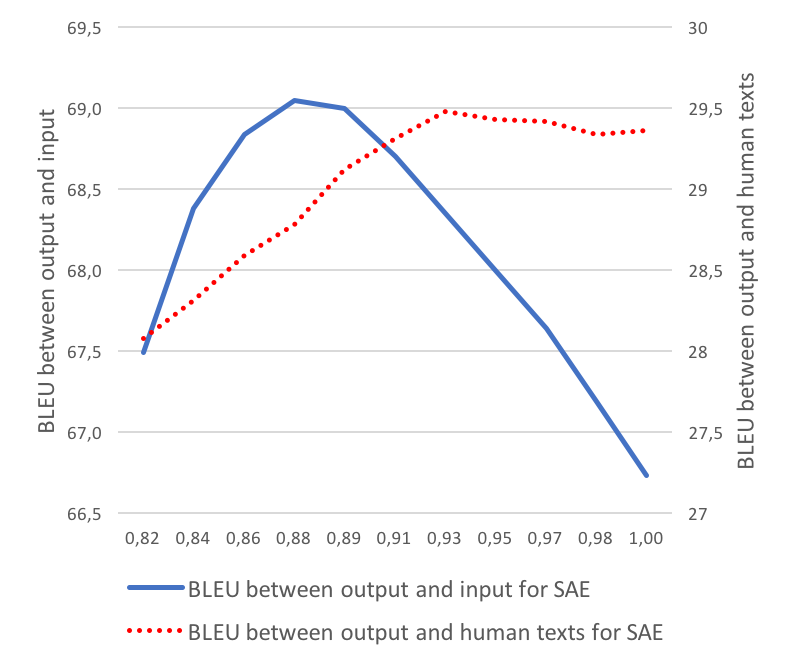}}
\caption{Manipulating the generated output in a way that boosts accuracy one can change BLEU between output and input. Moreover, such manipulation increases BLEU between output and human written reformulations. The picture shows behavior of SAE, but other architectures demonstrate similar behavior. The results are an average of four consecutive retrains of the same architecture.}
\label{pic:nakr}
\end{center}
\end{figure}

This way One can replace all sentences that push measured accuracy down and boost reported accuracy to 100\%. To see the effect that this manipulation has on the key performance metric we split all sentences with wrong style in 10 groups of equal size and replaces them with the best possible duplicates of the stylistically correct sentences group after group. The results of this process are shown in Figure \ref{pic:nakr}.

This result is disconcerting. Simply replacing part of the output with duplicates of the sentences that happen to have relatively high BLEU with given inputs allows to "boost" accuracy to 100\% and "improve" BLEU. The change of BLEU during such manipulation stays within error margins of the architecture, but accuracy is significantly manipulated. What is even more disturbing is that BLEU between such manipulated output of the batch and human-written reformulations provided in \cite{tian18} also grows. Figure \ref{pic:nakr} shows that for SAE but all four architectures described in Section \ref{sec:st} demonstrate similar behavior.

Our experiments show that though we can manipulate BLEU between output and human-written text, it tends to change monotonically. That might be because of the fact that this metric incorporates information on stylistics and semantics of the text at the same time, preserving inevitable entanglement that we have mentioned earlier. Despite being costly, human-written reformulations are needed for future experiments with style transfer. It seems that modern architectures have reached a certain level of complexity for which naive proxy metrics such as accuracy of an external classifier or BLEU between output and input are already not enough for performance estimation and should be combined with BLEU between output and human-written texts. As the quality of style transfer grows further one has to improve the human-written data sets: for example, one would like to have data sets similar to the ones used for machine translation with several reformulations of the same sentence. 

\begin{figure}[ht]
\begin{center}
\centerline{\includegraphics[width=\columnwidth]{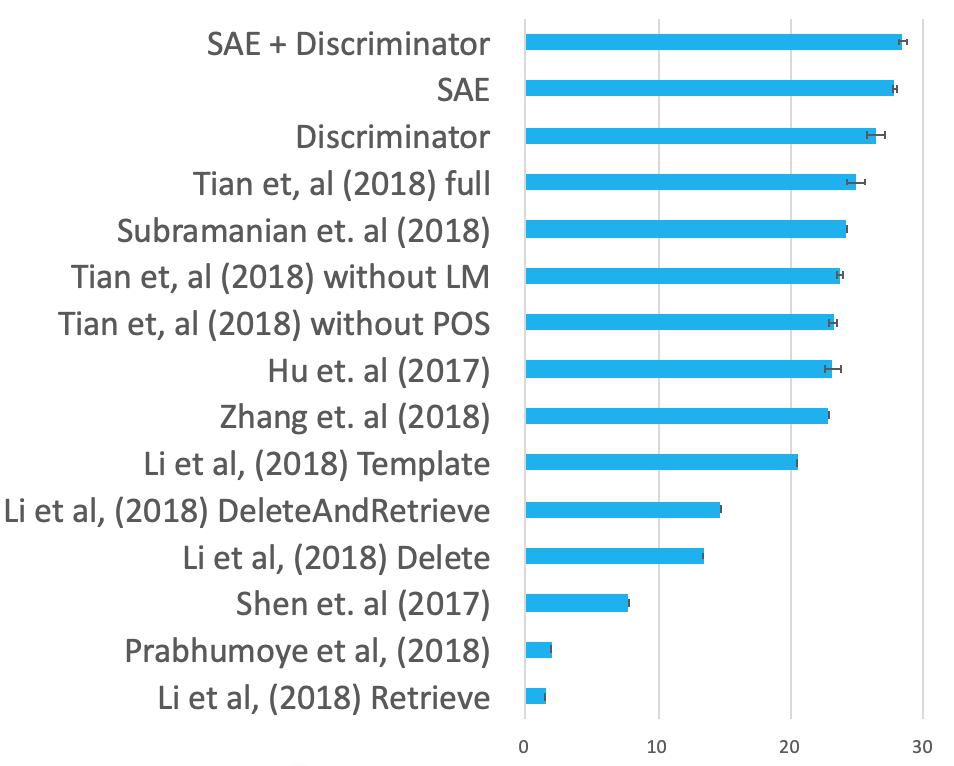}}
\caption{Overview of the BLEU between output and human-written reformulations of Yelp! reviews. Architecture with additional discriminator, shifted autoencoder (SAE) with additional cosine losses, and a combination of these two architectures measured after five re-runs outperform the baseline by \cite{hylsx} as well as other state of the art models. Results of \cite{romanov18} are not displayed due to the absence of self-reported BLEU scores}
\label{pic:ovreal}
\end{center}
\end{figure}

On Figure \ref{pic:ovreal} one can see how new proposed architectures compare with another state of the art approaches in terms of BLEU between output and human-written reformulations.

\section{Conclusion}

Style transfer is not a rigorously defined NLP problem. Starting from definitions of style and semantics and finishing with metrics that could be used to evaluate the performance of a proposed system. There is a surge of recent contributions that work on this problem. This paper highlights several issues connected with this lack of rigor. First, it shows that the state of the art algorithms are inherently noisy on the two most widely accepted metrics, namely, BLEU between input and output and accuracy of the external style classifier. This noise can be partially attributed to the adversarial components that are often used in the state of the art architectures and partly due to certain methodological inconsistencies in the assessment of the performance. Second, it shows that reporting error margins of several consecutive retrains for the same model is crucial for the comparison of different architectures, since error margins for some of the models overlap significantly. Finally, it demonstrates that even BLEU on human-written reformulations can be manipulated in a relatively simple way.

\bibliography{acl2019}
\bibliographystyle{acl_natbib}

\appendix

\section{Supplemental Material}
\label{sec:appendix}
Here are some examples characteristic for different systems. An output of a system follows the input. Here are some successful examples produced by the system with additional discriminator:

\begin{itemize}
    \item it's not much like an actual irish pub, which is depressing. $\rightarrow$ it's definitely much like an actual irish pub, which is grateful.
    \item i got a bagel breakfast sandwich and it was delicious! $\rightarrow$ i got a bagel breakfast sandwich and it was disgusting!
    \item i love their flavored coffee. $\rightarrow$ i dumb their flavored coffee.
    \item i got a bagel breakfast sandwich and it was delicious! $\rightarrow$  i got a bagel breakfast sandwich and it was disgusting!
    \item i love their flavored coffee. $\rightarrow$ i dumb their flavored coffee.
    \item nice selection of games to play. $\rightarrow$ typical selection of games to play.
    \item i'm not a fan of huge chain restaurants. $\rightarrow$ i'm definitely a fan of huge chain restaurants.
\end{itemize}

Here are some examples of typical faulty reformulations:
\begin{itemize}
\item only now i'm really hungry, and really pissed off. $\rightarrow$  kids now i'm really hungry, and really extraordinary off.
\item what a waste of my time and theirs. $\rightarrow$ what a wow. of my time and theirs.
\item cooked to perfection and very flavorful. $\rightarrow$ cooked to pain and very outdated.
\item the beer was nice and cold! $\rightarrow$ the beer was nice and consistant!
\item corn bread was also good! $\rightarrow$ corn bread was also unethical bagged
\end{itemize}

Here are some successful examples produced by the SAE:
\begin{itemize}
\item our waitress was the best, very accommodating. $\rightarrow$ our waitress was the worst, very accommodating.
\item great food and awesome service! $\rightarrow$ horrible food and nasty service!
\item their sandwiches were really tasty. $\rightarrow$ their sandwiches were really bland.
\item i highly recommend the ahi tuna. $\rightarrow$ i highly hated the ahi tuna.
\item other than that, it's great! $\rightarrow$ other than that, it's horrible!
\end{itemize}

Here are some examples of typical faulty reformulations by SAE:
\begin{itemize}
\item good drinks, and good company. $\rightarrow$ 9:30 drinks, and 9:30 company.
\item like it's been in a fridge for a week. $\rightarrow$ like it's been in a fridge for a true.
\item save your money \& your patience. $\rightarrow$ save your smile \& your patience.
\item no call, no nothing. $\rightarrow$ deliciously call, deliciously community.
\item sounds good doesn't it? $\rightarrow$ sounds good does keeps it talented
\end{itemize}

Here are some successful examples produced by the SAE with additional discriminator:
\begin{itemize}
\item best green corn tamales around. $\rightarrow$ worst green corn tamales around.
\item she did the most amazing job. $\rightarrow$ she did the most desperate job.
\item very friendly staff and manager. $\rightarrow$ very inconsistent staff and manager.
\item even the water tasted horrible. $\rightarrow$ even the water tasted great.
\item go here, you will love it. $\rightarrow$ go here, you will avoid it.
\end{itemize}

Here are some examples of typical faulty reformulations by the SAE with additional discriminator:
\begin{itemize}
\item \_num\_ - \_num\_ \% capacity at most , i was the only one in the pool. $\rightarrow$ sweetness - stylish \% fountains at most, i was the new one in the
\item this is pretty darn good pizza! $\rightarrow$ this is pretty darn unsafe pizza misleading
\item enjoyed the dolly a lot. $\rightarrow$ remove the shortage a lot.
\item so, it went in the trash. $\rightarrow$ so, it improved in the hooked.
\item they are so fresh and yummy. $\rightarrow$ they are so bland and yummy.
\end{itemize}

\end{document}